\documentclass[letterpaper, 10 pt, conference]{ieeeconf} 
\IEEEoverridecommandlockouts
\usepackage{printlen}

\usepackage[T1]{fontenc}
\usepackage{amsmath,amsfonts, amssymb}
\usepackage{cite}
\usepackage{stfloats}
\usepackage{graphicx}
\usepackage[caption=false,font=scriptsize]{subfig}
\usepackage{url}
\usepackage{hyperref}

\usepackage[per-mode = fraction]{siunitx}
\usepackage[nolist]{acronym}
\usepackage[none]{hyphenat}
\usepackage{glossaries}

\usepackage{bbm} 
\usepackage{algorithm}
\usepackage{algpseudocode}
\usepackage{booktabs}
\usepackage{tabularx}
\usepackage{multirow}
\usepackage{pifont}
\usepackage[table]{xcolor}

\usepackage{tikz, pgfplots, pgfplotstable}
\pgfplotsset{compat=1.18}
\usepgfplotslibrary{groupplots}
\usetikzlibrary{arrows, arrows.meta, patterns}
\usepgfplotslibrary{statistics}
\pgfplotsset{compat=1.18}
\usetikzlibrary{external}
\newcommand{\includetikz}[1]{%
    \includegraphics{figures/#1.pdf}
}

\usepackage{xcolor}
\definecolor{Black}{HTML}{000000}
\definecolor{Blue}{HTML}{0065bd}
\definecolor{Bluelight}{HTML}{D6E8F7}
\definecolor{Bluestrong}{HTML}{003359}
\definecolor{Red}{HTML}{8C000F}
\definecolor{Orange}{HTML}{E37222}
\definecolor{OrangePP}{HTML}{E97132}
\definecolor{Green}{HTML}{A2AD00}
\definecolor{GreenCR}{HTML}{008000}
\definecolor{LightGray}{HTML}{e7e7e7}
\definecolor{Gray}{HTML}{7f7f7f}
\definecolor{Gray-opac}{HTML}{d8d8d8}
\definecolor{MyDarkBlue}{RGB}{14,40,65}
\definecolor{TUMOrange}{RGB}{247,129,30}
\definecolor{TUMYellow}{RGB}{254,215,2}
\definecolor{TUMGreen}{RGB}{159,186,54}
\definecolor{TUMBlueDark4}{RGB}{20,81,154}
\definecolor{TUMBlueLightDark}{RGB}{154,188,228}

\definecolor{TUMBlueBrand}{HTML}{3070b3}
\definecolor{TUMBlueDark}{HTML}{072140}
\definecolor{TUMBlueDark5}{HTML}{165db1}
\definecolor{TUMBlueLight}{HTML}{5e94d4}
\definecolor{TUMBlueBright}{HTML}{8f81ea}
\definecolor{TUMRed}{HTML}{ea7237}
\definecolor{TUMRedDark}{RGB}{217,81,23}
\definecolor{TUMPink}{RGB}{181,92,165}
\definecolor{TUMBlueLight3}{HTML}{9ABCE4}
\definecolor{TUMGreyMid}{HTML}{A4ACB2}

\usepackage{textcomp}
\usepackage{lipsum}
\newcommand\copyrighttext{%
	\footnotesize \textcopyright 2026 IEEE.  Personal use of this material is permitted.  Permission from IEEE must be obtained for all other uses, in any current or future media, including reprinting/republishing this material for advertising or promotional purposes, creating new collective works, for resale or redistribution to servers or lists, or reuse of any copyrighted component of this work in other works.
}
\newcommand\copyrightnotice{%
	\tikzset{external/export=false}
	\begin{tikzpicture}[remember picture,overlay]
	\node[anchor=south,yshift=10pt, xshift=10pt] at (current page.south) {\fbox{\parbox{\dimexpr\textwidth-\fboxsep-\fboxrule\relax}{\copyrighttext}}};
	\end{tikzpicture}%
	\tikzset{external/export=true}
}

\graphicspath{{./figures/}}

\title{\LARGE \bf
Control-Informed Constraint Adaptation\\in Minimum-Time Trajectory Planning for Autonomous Racing
}

\newif\iffinal
\finaltrue

\newcommand{\authorsFinal}{
    \author{Ann-Kathrin Schwehn, Alexander Langmann, Mattia Piccinini and Johannes Betz%
    \thanks{A. Schwehn, A. Langmann, M. Piccinini and J. Betz are with the Professorship of Autonomous Vehicle Systems, TUM School of Engineering and Design, Technical University of Munich, 85748 Garching, Germany; Munich Institute of Robotics and Machine Intelligence (MIRMI).
    Contact: \{ann-kathrin.schwehn, alexander.langmann, mattia.piccinini, johannes.betz\} @tum.de
    }%
    }
}

\newcommand{\authorsBlind}{%
  \author{Anonymous Author(s)%
  \thanks{This work has been submitted to ITSC 2026 for peer review. Affiliations of the author(s) are omitted for double-blind review.}%
  }
}

\iffinal
  \authorsFinal
\else
  \authorsBlind
\fi

\begin{document}
\bstctlcite{BSTcontrol}

\maketitle
\copyrightnotice

\begin{abstract}

  Autonomous racecars operate at the limits of vehicle dynamics, where small control errors translate into safety-critical behavior and lost performance. Trajectory planners assume perfect tracking and remain blind to execution errors. To guarantee safety, trajectory planners therefore restrict themselves to conservative spatial margins, leaving usable track space untapped.
  To overcome these issues, we introduce a control-informed online trajectory planning framework that learns from its own execution errors. By measuring systematic tracking deviations during runtime, we dynamically adapt spatial track constraints and iteratively expand the free-space planning area. The planner remains time-optimal while compensating for accumulated execution errors.
  This method was analyzed in a high-fidelity closed-loop simulation environment with autonomous racecars. The results demonstrate that our approach reduces lap time by 1.8\,s without increasing computational burden, maintaining a median runtime of 25\,ms.
  Our finding indicates that feeding control-induced deviations back into the planning layer unlocks performance previously inaccessible to modular architectures and enables autonomous vehicles to exploit track limits systematically.
\end{abstract}

\section{Introduction}
\label{sec:introduction}

Executing time-optimal maneuvers near the handling limits remains a core challenge in autonomous racing \cite{Betz.2022}. Under these conditions, highly nonlinear vehicle dynamics and inevitable model mismatches inherently lead to control deviations. While global trajectory planners excel at generating curvature- \cite{Heilmeier.2020} or time-optimal \cite{Lovato.2022} racelines, they generally do so without considering the actual tracking capabilities of the downstream controller. Existing optimization-based local planners can generate online time-optimal maneuvers \cite{Rowold.2023,Piccinini.2024a,Subosits.2019}, but do not dynamically adapt their operation based on observed execution errors. As a result, these planners assume perfect execution while the control modules physically struggle to maintain the planned trajectory. Because of this structural separation, the theoretical performance limits of offline time-optimal racelines are not fully exploited during online execution \cite{Kabzan.2019} because the planner has no information about the global system's behavior. To handle closed-loop execution errors and ensure the vehicle stays within the physical track boundaries, most existing software stacks rely on safety margins. This creates a functional distinction between the \textit{drivable area} (the physical track surface) and the \textit{plannable area} (the specific spatial region where a trajectory can be generated). In conventional approaches, the plannable area is bounded by static, often conservative margins. Since existing time-optimal motion planners lack a feedback mechanism to actively adapt their plannable area in the presence of execution errors, their trajectories are permanently confined to fixed spaces. Instead, feedback from the control to the planning module can prevent overly conservative behavior, maximize track utilization and ultimately enable better laptimes. To overcome these limitations, we propose an online-capable, control-informed trajectory planning framework that establishes a direct feedback loop from the control module back to the planner. By continuously observing the execution error of the system during runtime, our method feeds this execution data back to the planning module, which dynamically adjusts the plannable area accordingly. By feeding these updated constraints into a Model Predictive Control (MPC) planner, we iteratively refine the plannable area to actively compensate for systematic deviations. This iterative refinement significantly improves utilization of the drivable area, allowing the system to push closer to the vehicle's true physical limits and to enhance the overall system behavior.

\begin{figure}[t]
    \vspace*{4.5mm}
    \centering
    \resizebox{\linewidth}{!}{\includetikz{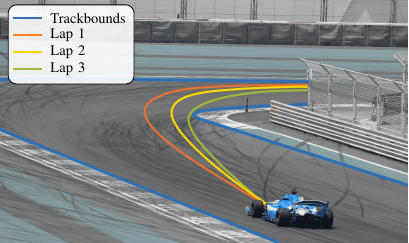}}
    \vspace{-7mm}
    \caption{Progression of the driven trajectory over consecutive laps. By iteratively adapting the plannable area based on closed-loop execution errors, the proposed framework enables the vehicle to approach the physical track boundaries and utilize previously unexploited track space.}
    \label{fig:first_image}
\end{figure}

The paper is structured as follows: Section \ref{sec:relatedwork} reviews related literature and outlines our contributions. Section \ref{sec:method} details the system architecture and the mathematics behind the adaptive constraint formulation. Section \ref{sec:results} presents the simulation setup and discusses the results, followed by concluding remarks in Section \ref{sec:conclusion}.
\section{Related Work}
\label{sec:relatedwork}

In autonomous racing software stacks, global trajectory planning addresses the generation of an optimal raceline based on track features and vehicle capabilities \cite{Betz.2022}. This task is often performed offline using complex vehicle models to achieve high fidelity, as strict real-time computational constraints do not apply \cite{Subosits.2019}. Various approaches decouple this problem into sequential path and velocity generation \cite{Heilmeier.2020, Velenis.2008, Subosits.2015, Kapania.2016}. For instance, Heilmeier et al. \cite{Heilmeier.2020} generate a minimum curvature path in 2D and subsequently compute the velocity profile using a forward-backward solver. Coupled approaches jointly optimize the path and velocity profile by formulating a Minimum Lap Time Problem (MLTP) and solving it directly via an Optimal Control Problem (OCP). These coupled formulations are applied to both 2D \cite{Veneri.2020, Rucco.2015, Christ.2021} and 3D track representations \cite{Limebeer.2015, Lovato.2022, vandenEshof.2025}. However, offline methods inherently lack the capability to react to dynamic environmental changes or state deviations during runtime.

\begin{figure*}[t]
    \centering
    \includegraphics[width=0.9\textwidth, trim=0 1.0cm 0 0, clip]{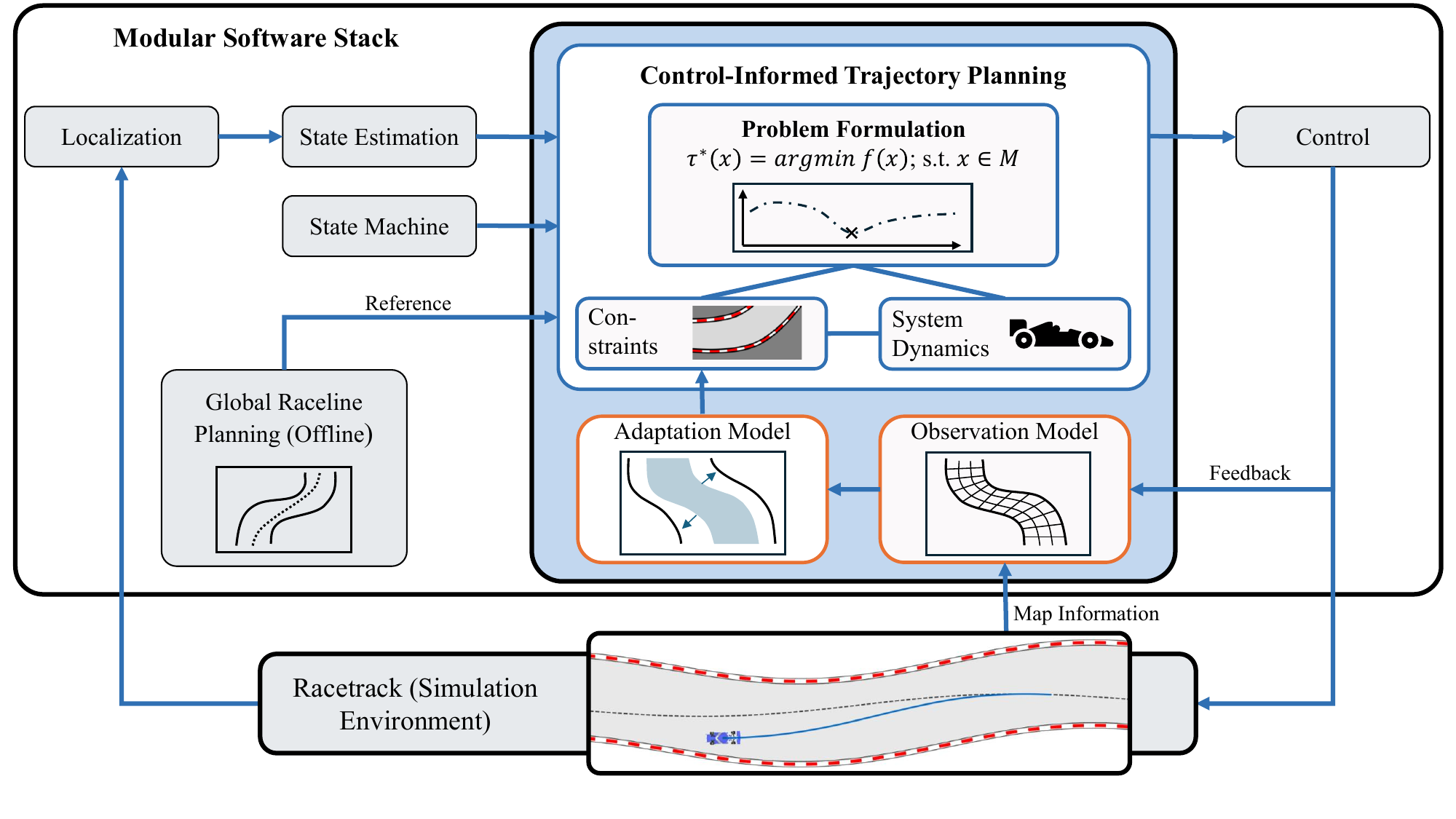}
    \caption{System architecture of the proposed control-informed trajectory planning framework integrated into a modular autonomous racing software stack. A direct feedback loop connects the downstream controller to the planning module, where the observation model evaluates execution deviations and the adaptation model iteratively updates the spatial constraints of the OCP.}
    \label{fig:overview}
\end{figure*}

To address this limitation, the MLTP can be adapted for online application by reducing the planning horizon, transforming the problem into a receding-horizon optimization task. Optimization-based online planners continuously recalculate the optimal trajectory to account for the current vehicle state \cite{Gundlach.2019, Rowold.2023, Pagot.2020}. Recent literature has focused on refining these online formulations to enable more diverse maneuvers, either by varying acceleration constraints \cite{Piccinini.2024b} or adjusting cost function designs \cite{Taddei.2025}. Despite these improvements, online planners typically assume idealized trajectory execution and remain largely unaware of systematic tracking errors that inevitably occur at absolute handling limits.

To compensate for these deviations, online adaptations are classically confined to the control module. Given the highly repetitive nature of closed-circuit racing, iterative learning-based control strategies \cite{Rosolia.2018, Kabzan.2019, Kapania.2020}, Gaussian Process-based tracking MPC \cite{Hewing.2018}, and data-driven stochastic tracking \cite{Carrau.2016, Zeng.2025} are frequently employed to improve lateral and longitudinal tracking over consecutive laps. Additionally, Jahncke et al. \cite{Jahncke.2026} adjust tracking MPC weights dynamically according to current driving behavior. While effective at stabilizing the vehicle locally, these approaches only feed deviation data back into the control module, leaving the upstream planning module entirely uninformed.

Combining idealized global planning and actual vehicle control requires closed-loop-aware architectures. Gulisano et al. \cite{Gulisano.2025} deploy a stochastic vehicle model to propagate covariance along a planned trajectory, but fail to perform the planning task online. Acting as a feedback loop from control to planning, Werner et al. \cite{Werner.2025} provide spatially resolved acceleration limit adaptations to a planner based on vehicle state observations from prior runs. However, this adaptation is strictly an offline procedure requiring expert tuning. Wachter et al. \cite{Wachter.2026} adjust acceleration constraints during runtime and perform trajectory adaptation, but do not address compensation of lateral errors and track utilization.

From the analysis of the existing literature, we derive that there is a lack of a systematic, online planning method capable of dynamically adjusting spatial track constraints to compensate for accumulated control errors. To address this research gap, we propose the following three contributions:
\begin{itemize}
    \item We extend an online time-optimal MPC trajectory planner to generate execution error-aware maneuvers for autonomous racecars.
    \item We introduce a new method to explicitly evaluate control-induced, closed-loop execution errors during runtime, and proactively compensate for these offsets by adapting the plannable area constraints in our MPC planner.
    \item We provide a closed-loop simulation study in a high-fidelity racing simulator, demonstrating that our constraint-adaptive planner improves overall system performance compared to a non-adaptive baseline.
\end{itemize}

\section{Methodology}
\label{sec:method}

In this section, we introduce the proposed control-informed trajectory planning framework, designed to dynamically adapt the plannable area during runtime. Our algorithm is integrated into a modular autonomous racing software stack \cite{Hoffmann.2026}. We further present the track representation, the vehicle dynamics, and the baseline OCP. Finally, we display our core contribution in detail, presenting the mathematical formulations used to evaluate tracking errors and iteratively adapt the plannable area constraints.

\subsection{System Architecture}
\label{subsec:architecture}

Our approach is applied to the challenging domain of full-scale autonomous racing at the dynamical limits. We integrate our control-informed trajectory planning algorithm into a modular autonomous racing software stack, as illustrated in Figure \ref{fig:overview}.
Our online trajectory planner receives data from three upstream modules. The state estimation module provides the current vehicle state vector, including spatial position, velocity, acceleration, heading, and yaw rate. A globally optimal raceline computed offline based on \cite{Rowold.2023} acts as the main spatial reference. Additionally, a state machine provides operational commands such as target speeds or stop requests.

Operating in a closed control loop, the planner generates a target trajectory and forwards it to the downstream tracking controller. To establish the feedback mechanism, an observation module evaluates the spatial error of the physically executed trajectory against the optimal reference. This tracking deviation data is passed directly to an adaptation module, which recalculates the spatial constraints of the plannable area based on the mechanisms presented in Section \ref{subsec:constraints}. These updated boundary constraints are spatially stored along the track and applied at the corresponding locations during the subsequent lap, enabling the system to iteratively compensate for downstream execution errors.

\subsection{Online Time-Optimal Trajectory Planning}
\label{subsec:problem}
Our baseline trajectory planner solves online an optimization-based MLTP. Its objective is to plan new time-optimal maneuvers from the current vehicle state, without tracking the global raceline.

We adopt a ribbon-based, spatially-parameterized curve model as proposed in \cite{Perantoni.2014} to represent the track in 3D and achieve better vehicle modeling. We utilize a Frenet coordinate frame $(s, n)$ to describe the vehicle's progress $s$ along the reference line and its lateral deviation $n$.

To ensure computational efficiency for online application, the vehicle is modeled as a 3D point mass, following the formulations from \cite{Rowold.2023}. The system state vector is defined as:
\begin{equation}
    \mathbf{x} = \begin{bmatrix}
        V & n & \hat{\chi} & \hat{a}_\mathrm{x} & \hat{a}_\mathrm{y}
    \end{bmatrix}^{\top},
\end{equation}
where $V$ is the total velocity, $\hat{\chi}$ is the relative orientation of the velocity vector, and $\hat{a}_\mathrm{x}, \hat{a}\mathrm{y}$ are the longitudinal and lateral accelerations, respectively. The longitudinal and lateral jerks serve as control inputs $\mathbf{u} = [\hat{j}_\mathrm{x}, \hat{j}_\mathrm{y}]^{\top}$. The continuous-time spatial system dynamics are denoted as $\mathbf{x}' = \frac{1}{\dot{s}}\mathbf{f}(\mathbf{x}, \mathbf{u})$, with the full derivation provided in \cite{Rowold.2023}. We employ a standard g-g-g-v formulation to enforce three-dimensional acceleration limits \cite{Werner.2025}.

The online planner solves a receding-horizon MLTP formulated (similar to \cite{Rowold.2023}) as:
\begin{subequations}
    \label{eq:OCP}
    \begin{align}
        \min_{\mathbf{x},\mathbf{u}} \!\!
         & \int_{s_0}^{s_\mathrm{e}}\left(
        \frac{1}{\dot{s}}
        + \mathbf{u}^T \mathbf{R}\,\mathbf{u} + \begin{bmatrix}
                                                    1 \\
                                                    \epsilon
                                                \end{bmatrix}^{\top}
        \mathbf{S} \begin{bmatrix}
                       1 \\
                       \epsilon
                   \end{bmatrix}
        \right)\,ds \label{eq:costfunction}                                                                                                       \\
        \text{s.t.}\quad
         & \mathbf{x}' = \frac{1}{\dot{s}}\,\mathbf{f}(\mathbf{x},\mathbf{u}), \label{eq:pointmass}                                               \\[3pt]
         & g\bigl(V,a_\mathrm{x},a_\mathrm{y},a_\mathrm{z})\,\le\,0, \label{eq:gggv}                                                              \\
         & V - \epsilon \le V_{s_\mathrm{max}}, \label{eq:V_eps}                                                                                  \\
         & n_\mathrm{r}(s) + d_\mathrm{s} + \Delta b_{\mathrm{r}} \le n \le n_\mathrm{l}(s) - d_\mathrm{s} + \Delta b_{\mathrm{l}}, \label{eq:tb} \\
         & -\tfrac{\pi}{2}\le\hat{\chi}\le\tfrac{\pi}{2}. \label{eq:heading}
    \end{align}
\end{subequations}
In \eqref{eq:costfunction}, $s_0$ and $s_\mathrm{e}$ denote the longitudinal start and end coordinates of the current planning horizon, respectively. The first term in \eqref{eq:costfunction} enforces the minimization of the maneuver time along the planning horizon. Furthermore, the matrix $\mathbf{R}$ regularizes jerk, while $\mathbf{S}$ prevents velocity overshoots. The slack variable $\epsilon$ allows setting speed caps, so $V_{\mathrm{max}} \le V(s_0)$ becomes feasible. The dynamics of the vehicle model are included in \eqref{eq:pointmass} with a conversion in the spatial domain. Constraint \eqref{eq:gggv} enforces the accelerations to be within the vehicle's limits, while the inequality constraint \eqref{eq:V_eps} enables speedcaps for the autonomous racecar. In \eqref{eq:tb}, it is ensured that the planned trajectory lies within the plannable area, where $n_\mathrm{r, l}(s)$ are the track widths from the reference line to the left or right track bound, $d_\mathrm{s}$ is a static safety margin and $\Delta b_{\mathrm{r}}$ dynamically shifts the safety margin, further explained in Section~\ref{subsec:constraints}. Finally, constraint \eqref{eq:heading}
ensures that the car drives in the correct direction on the track.

\subsection{Adaptive Constraints}
\label{subsec:constraints}
\begin{figure}[t]
    \centering
    \resizebox{\linewidth}{!}{\includetikz{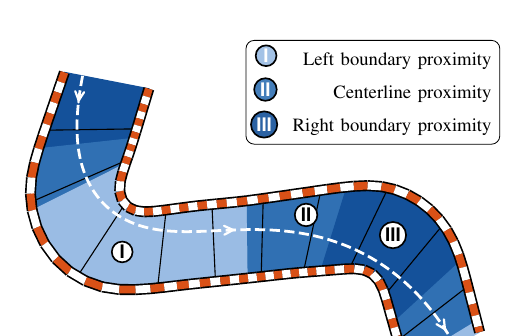}}
    \caption{Depiction of the segmentation of the race track. The sections shaded in blue indicate the proximity of the raceline to a track boundary. The thin black lines mark the discrete points of the adaptation map.}
    \label{fig:track_sections}
\end{figure}

To improve tracking of an offline-computed ideal raceline during runtime, we iteratively adapt the plannable area depending on the lateral offset of the actual driven trajectory to the raceline. This raceline deviation is determined by an observation model as:
\begin{equation}
    \Delta d_{i,j} = d_{\mathrm{rl},j} - d_{\mathrm{driven},i,j},
    \label{eq:deviation}
\end{equation}
where $d_\mathrm{rl,j}$ is the lateral position of the offline-computed raceline, $d_{\mathrm{driven},i,j}$ is the actual driven trajectory, $i$ denotes the current lap index, and $j$ denotes the spatial grid cell index in an \textit{adaptation map}. In this map, the boundary adaptations of lap $i$ are stored at spatial location $j$, utilizing discrete points with a longitudinal distance of \SI{1}{\meter} in between in this work.

The observation model determines which side of the track boundary requires a safety margin adaptation to minimize $\Delta d_{i,j}$. As illustrated in \autoref{fig:track_sections}, we divide the race track into \textit{sections} of three primary spatial relationships between an ideal line and the track bounds. \textit{Sections~I} and \textit{III} depict situations in which the ideal path closely approaches a specific track bound. Specifically, \textit{section~I} marks areas close to the left bounds, i.e., the apex of a left turn or the entry/exit of a right turn. The contrary principle governs \textit{section~III} with areas close to the right bounds and therefore apexes of right turns, as well as corner entries and exits of a left turn. In both \textit{sections}, the respective safety margin on the side closer to the track boundary is adapted to influence the planner's behavior. Finally, in \textit{section~II}, the optimal line does not approach either boundary. Therefore, safety margins on both sides are symmetrically adjusted.

To enable the planning algorithm to dynamically enlarge or reduce the plannable area, we introduce four heuristics to determine the safety margin adaptation $\Delta b_{i,j}$. Boundary updates only trigger when the deviation from the previous lap $\Delta d_{i-1,j}$ exceeds a predefined threshold $\tau$. For all heuristics, we assume that the execution error is primarily deterministic and spatially dependent. Based on this assumption, all heuristics use iteration-domain filtering to capture the system behavior lap by lap. Further, two approaches include spatial information and correlation to prevent discontinuities in the constraints. This retains the stability and feasibility of the motion planner.

\textbf{Linear Heuristic H1:} The baseline model accumulates the scaled deviation over previous laps as:
\begin{equation}
    \Delta b_{i,j} = \begin{cases}
        \sum_{k=1}^{i-1} \lambda \Delta d_{k,j} & \text{if } \Delta d_{i-1,j} > \tau, \\
        0                                       & \text{otherwise},
    \end{cases}
    \label{eq:v1}
\end{equation}
where $\Delta b_{i,j}$ is the resulting shift applied to the safety margin constraint in lap $i$ at the spatial location $j$. The scaling factor $\lambda$ determines the strategy's aggressiveness. Lower values lead to plannable area adjustments of smaller magnitude, while higher values enable more aggressive adaptation. In all proposed approaches, the scaling factor $\lambda$ can be configured statically prior to a run or decay exponentially as the lap count increases. These dynamically updated values are stored in the adaptation map to serve as a look-up for the following laps.

\textbf{Curvature-scaled Heuristic H2:} The curvature-scaled adaptation incorporates the normalized curvature of the offline raceline to the baseline heuristic. The absolute curvature $\kappa$ is normalized to $\tilde{\kappa} \in [0,1]$ using:
\begin{equation}
    \tilde{\kappa} = \frac{\kappa - \kappa_{\mathrm{min}}}{\kappa_{\mathrm{max}}-\kappa_\mathrm{{min}}}.
    \label{eq:normalize}
\end{equation}
Because the change in curvature difference between the offline and online raceline is negligible in our setting, $\tilde{\kappa}_{j}$ is precomputed and stored in the adaptation map, avoiding online computation overhead. The boundary shift is defined as:
\begin{equation}
    \Delta b_{i,j} =
    \begin{cases}
        \sum_{k=1}^{i-1} \lambda \tilde{\kappa}_{j} \Delta d_{k,j} & \text{if } \Delta d_{i-1,j} > \tau, \\
        0                                                          & \text{otherwise}.
    \end{cases}
    \label{eq:v2}
\end{equation}

\textbf{Low-pass Filter H3:} A low-pass filter can be applied to update the constraints based primarily on the immediately preceding lap. This approach smooths the adaptation iteratively using:

\vspace{-3mm}
\begin{small}
    \begin{equation}
        \Delta b_{i,j} =
        \begin{cases}
            \lambda \Delta b_{i-1,j} + (1-\lambda) \Delta d_{i-1,j} & \text{if } \Delta d_{i-1,j} > \tau, \\
            0                                                       & \text{otherwise}.
        \end{cases}
        \label{eq:v3}
    \end{equation}
\end{small}

\textbf{Gaussian Heuristic H4:}
To account for the trajectory behavior in the immediate vicinity of the current position, the low-pass filter can be extended to incorporate the deviations of $p$ surrounding grid cells. We first define the spatially smoothed deviation $\tilde{d}_{i-1,j}$ as:
\begin{equation}
    \tilde{d}_{i-1,j} = \sum_{m=-p}^p w_m \Delta d_{i-1,j+m},
\end{equation}
where $w_m$ is a Gaussian weight distribution ensuring that cells closer to the current cell $j$ exert a stronger influence. The constraint shift is then computed as:

\vspace{-3mm}
\begin{small}
    \begin{equation}
        \Delta b_{i,j} =
        \begin{cases}
            \lambda \Delta b_{i-1,j} + (1 - \lambda) \tilde{d}_{i-1,j} & \text{if } \Delta d_{i-1,j} > \tau, \\
            0                                                          & \text{otherwise}.
        \end{cases}
        \label{eq:v4}
    \end{equation}
\end{small}
Here, the local deviation is smoothed using a Gaussian weight distribution $w_m$, ensuring that cells closer to the current cell $j$ exert a stronger influence on $\Delta b_{i,j}$.

\section{Results \& Discussion}
\label{sec:results}

\subsection{Experimental Setup}
\label{subsec:setup}
To evaluate the proposed control-informed planning framework, all experiments are conducted in a high-fidelity autonomous racing simulation environment derived from real-world data. The environment operates in a closed-loop configuration, utilizing an MPC tracking controller downstream of the planner. The offline-computed global raceline, which serves as the spatial reference for the observation module, is generated based on \cite{Rowold.2023}.

To systematically evaluate the proposed methodologies, the framework is validated on the Yas Marina Circuit, with each test run spanning five complete laps. All simulations are executed on a workstation equipped with an AMD Ryzen 7 PRO 7840U CPU and 32 GB of RAM.

\begin{figure*}[t]
    \centering
    \resizebox{0.9\linewidth}{!}{
        \includetikz{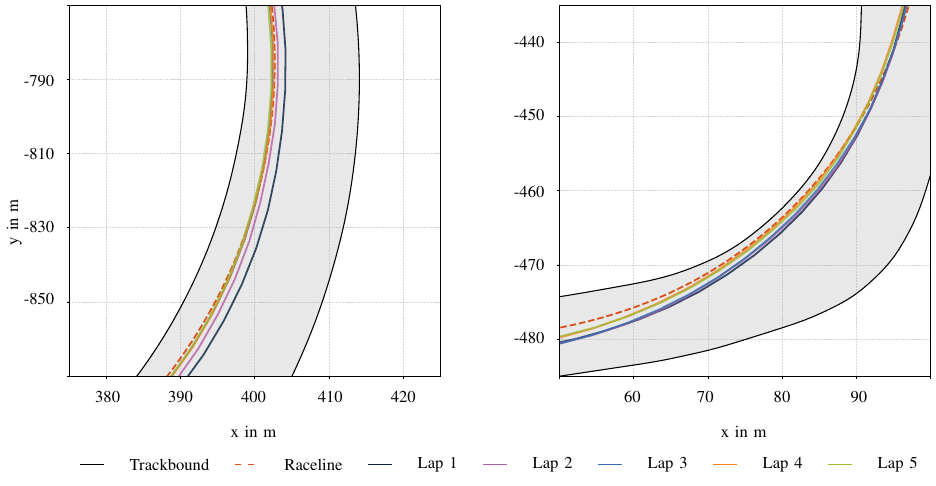}}
    \vspace{-4mm}
    \caption{Spatial progression of the driven trajectory over consecutive laps in turn 9 (left) and turn 13 (right). The adaptive constraints allow the vehicle to iteratively minimize the lateral error to the optimal raceline.}
    \label{fig:progression}
\end{figure*}

\subsection{Plannable Area Adaptation}
\label{subsec:plannable_area}
\begin{figure}[h]
    \centering
    \resizebox{\linewidth}{!}{
        \includetikz{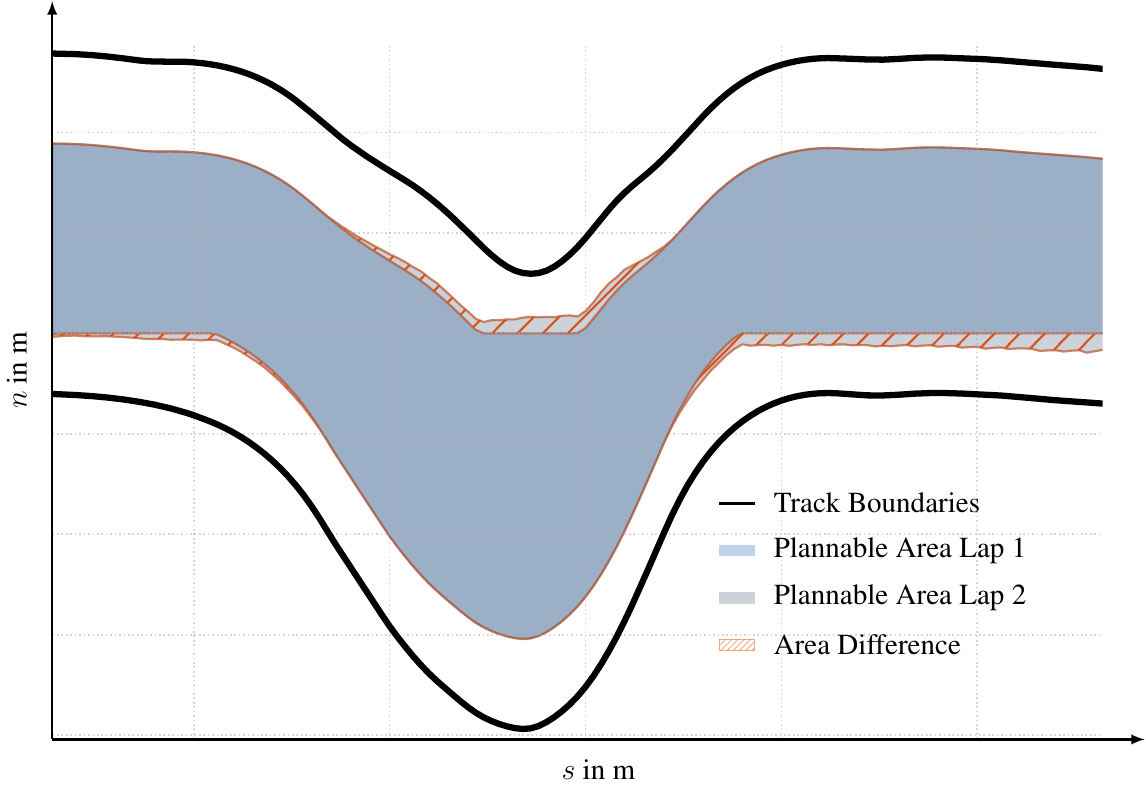}}
    \vspace{-5mm}
    \caption{Evolution of the plannable area using the Gaussian Heuristic (H4), depicted in the Frenet Frame (centered around the raceline). The hatched region illustrates the additional spatial margin gained through the iterative adaptation process compared to the area of the first lap.}
    \label{fig:plannable_area}
\end{figure}

To illustrate the active mechanism of the control-informed feedback loop, Figure \ref{fig:plannable_area} depicts the cross-sectional evolution of the plannable area within turn 5 using the Gaussian Heuristic (\textbf{H4}). During the initial lap, the plannable area is confined by conservative baseline margins and accounts for deviations observed in the window behind the vehicle. As tracking deviations are evaluated during runtime, the adaptation module iteratively shifts these boundary constraints. The hatched region highlights the spatial difference generated by this constraint shift. This expansion effectively increases the available maneuvering space, allowing the optimization problem to yield trajectories that utilize regions much closer to the physical track limits without violating safety constraints.

\subsection{Spatial Path Evolution}
\label{subsec:path_evolution}
Over the five-lap horizon, the executed trajectories exhibit strong spatial convergence. Utilizing (\textbf{H1}), Figure \ref{fig:progression} details this trajectory evolution across two distinct track segments. In turn 9, the initial executed trajectory exhibits a lateral deviation exceeding \SI{2}{\meter} relative to the optimal raceline. By the second lap, this error is reduced by approximately 50\%. Subsequent laps display asymptotic improvement. The lateral deviation is iteratively minimized, though a minor overshoot remains observable at the corner exit. The behavior in turn 13 (right plot in Figure \ref{fig:progression}) further demonstrates our method's capability to iteratively shift the driven path closer to the ideal spatial reference. Table \ref{tab:RMSE} reports the Root Mean Squared Error (RMSE) of the lateral deviation from the raceline over consecutive laps: our (\textbf{H1}) adaptation strategy reduces the deviations by 52.5\% in 5 laps.

\begin{table}[htbp]
    \centering
    \caption{RMSE of the lateral raceline deviation with closed-loop planning and control using our (H1) adaptation strategy.}
    \label{tab:RMSE}
    \small
    \begin{tabular}{lcc}
        \toprule
        \textbf{Laps} & \textbf{RMSE in m} \\
        \midrule
        1             & $0.8617$           \\
        2             & $0.5802$           \\
        3             & $0.4832$           \\
        4             & $0.4174$           \\
        5             & $\textbf{0.4089}$  \\
        \bottomrule
    \end{tabular}
\end{table}

\subsection{Sensitivity Analysis}
\label{subsec:sensitivity}

To systematically evaluate the proposed methodologies, the scaling factor $\lambda$ is designated as the primary independent variable across all four heuristics. Additionally, for the Gaussian approach (\textbf{H4}), the standard deviation $\sigma$ is systematically varied. An overview of the tested parameter combinations and their corresponding best lap times is detailed in Table \ref{tab:testcases}. Our results show that (\textbf{H1}) performed the best overall, while the other heuristics produced consistent results that are at least \SI{0.5}{\second} slower than (\textbf{H1}).

\begin{table}[ht]
    \centering
    \caption{Hyperparameter configurations and resulting best lap times for the evaluated constraint adaptation heuristics.}
    \label{tab:testcases}
    \begin{tabular}{l c c c c}
        \toprule[1.5pt]
        \textbf{Heuristic}             & \textbf{$\mathbf{\lambda}$} & \textbf{Best lap in s} & \textbf{$p$} & \textbf{$\sigma$} \\
        \midrule
        Linear (\textbf{H1})           & 1.0                         & \textbf{114.91}        & -            & -                 \\
                                       & 0.5                         & 115.21                 & -            & -                 \\
        \midrule
        Curvature-scaled (\textbf{H2}) & 1.0                         & 115.65                 & -            & -                 \\
                                       & 0.5                         & 115.65                 & -            & -                 \\
        \midrule
        Low-pass Filter (\textbf{H3})  & 0.5                         & 115.64                 & -            & -                 \\
                                       & 0.7                         & 115.66                 & -            & -                 \\
                                       & 0.3                         & 115.50                 & -            & -                 \\
        \midrule
        Gaussian (\textbf{H4})         & 0.7                         & 115.50                 & 8            & 2.5               \\
                                       & 0.5                         & 115.65                 & 8            & 2.5               \\
                                       & 0.3                         & 115.65                 & 8            & 2.5               \\
                                       & 0.7                         & 115.80                 & 8            & 1.5               \\
                                       & 0.5                         & 115.64                 & 8            & 1.5               \\
                                       & 0.3                         & 115.64                 & 8            & 1.5               \\
                                       & 0.7                         & 115.81                 & 8            & 4.0               \\
                                       & 0.5                         & 115.63                 & 8            & 4.0               \\
                                       & 0.3                         & 115.51                 & 8            & 4.0               \\
        \bottomrule[1.5pt]
    \end{tabular}
\end{table}

\subsection{Lap Time Progression}
\label{subsec:laptime_progression}

\begin{figure}[h]
    \centering
    \resizebox{\linewidth}{!}{
        \includetikz{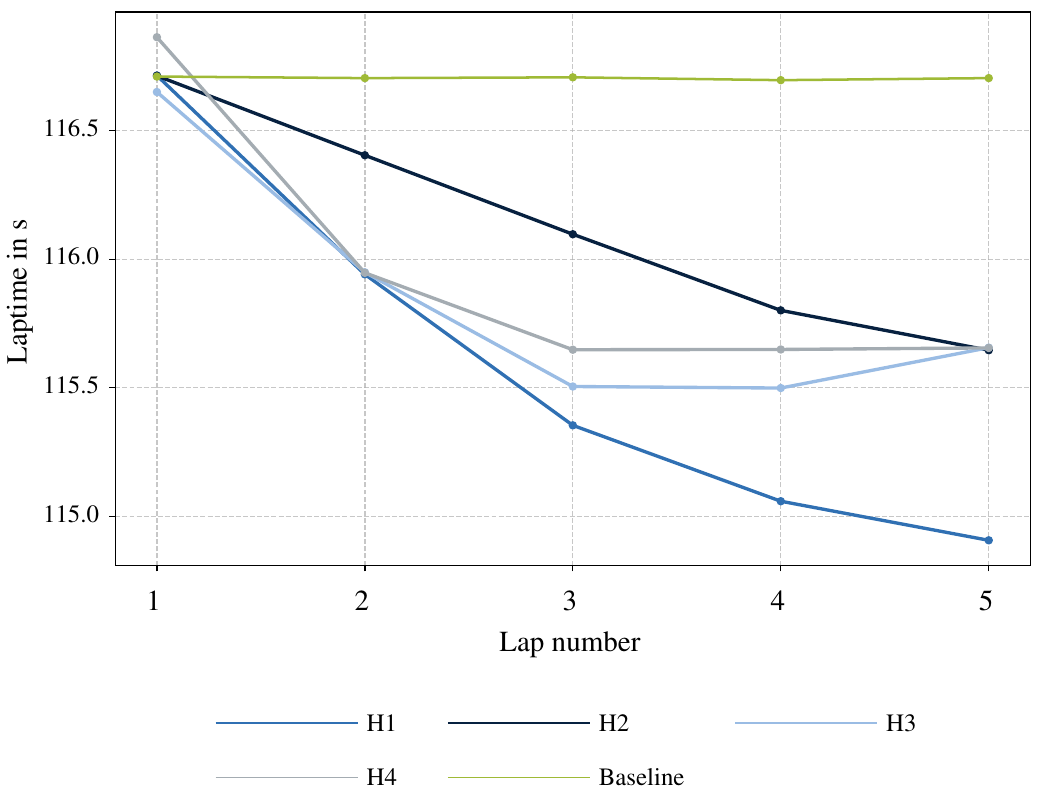}}
    \vspace{-8mm}
    \caption{Lap time progression of different heuristics and the non-adaptive baseline. The Linear Heuristic (H1) achieves the lowest overall lap time.}
    \label{fig:laptimes}
\end{figure}

Using the optimal hyperparameter configurations identified in Section \ref{subsec:sensitivity} for each heuristic, Figure \ref{fig:laptimes} illustrates the lap time progression of the evaluated heuristics over five consecutive laps. The non-adaptive baseline planner maintains a constant lap time of \SI{116.70}{\second} over multiple laps. Minor initial variations observed in the first lap across all adaptive methods are expected and attributed to noise added to the state estimation derived from real-world data.

The Linear Heuristic (\textbf{H1}) obtains the fastest lap time improvement and achieves the lowest overall lap time of \SI{114.91}{\second} by the fifth lap. This represents a performance gain of approximately \SI{1.8}{\second} compared to the static baseline.

The remaining heuristics show distinct adaptation characteristics. The Gaussian Heuristic (\textbf{H4}) converges rapidly after the second lap but plateaus early, effectively finding a stable but suboptimal limit. The Curvature-scaled Heuristic (\textbf{H2}) displays a consistent but notably slower convergence rate, requiring more laps to safely push the vehicle toward the physical track boundaries. In contrast, the Low-pass Filter (\textbf{H3}) achieves strong initial lap time reductions but experiences a slight performance degradation in the final lap. This regression highlights an overcompensation effect, where the iteratively adapted constraints become overly aggressive and cause the downstream tracking controller to struggle.

The lap time results indicate that all proposed variants of the control-informed feedback loop successfully improve track utilization and outperform the static baseline to varying degrees.

\subsection{Runtime Analysis}
\label{subsec:runtime}
\begin{figure}[h]
    \centering
    \resizebox{\linewidth}{!}{
        \includetikz{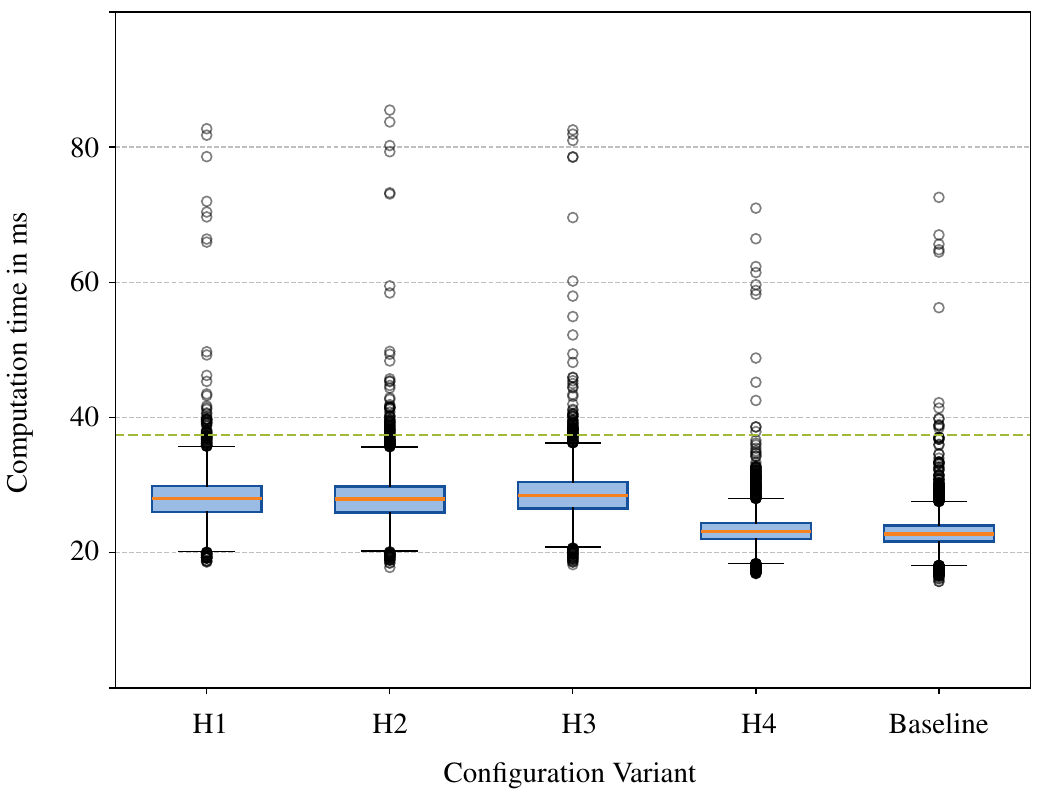}}
    \vspace{-7mm}
    \caption{Distribution of computation times across the configuration variants.}
    \label{fig:comp_times}
\end{figure}

To investigate the online capability of our framework, we benchmark the execution times of the presented heuristics \textbf{H1-H4} against the non-adaptive baseline. Figure \ref{fig:comp_times} illustrates the distribution of computation times across the different configuration variants.

The integration of the spatial observation module and the constraint adaptation map introduces minimal computational overhead. Across all tested heuristics, the median computation time ranges consistently between \SI{25}{\milli\second} and \SI{30}{\milli\second}. Furthermore, the maximum computation times remain strictly below \SI{100}{\milli\second} in all test cases. This confirms that dynamically updating the plannable area based on control feedback does not compromise the online viability of the MPC planner.

\subsection{Discussion}
\label{subsec:discussion}
Our findings indicate iteratively updating the plannable area enables the vehicle to utilize track regions inaccessible to statically constrained planners. The continuous lap time reductions suggest a correlation with the lateral tracking error, as shown by the halving of the RMSE over five laps for \textbf{H1}. The runtime analysis confirms this spatial adaptation adds negligible computational overhead, making it suitable for online application. While all evaluated heuristics expand the boundaries, their convergence behaviors indicate a clear trade-off. Aggressive error accumulation (\textbf{H1}) maximizes lap time improvements, whereas temporal filtering (\textbf{H3}) appears susceptible to overcompensation, generating abrupt constraint gradients that the tracking controller struggles to follow. To mitigate this, the Gaussian heuristic (\textbf{H4}) distributes deviations across neighboring cells, which yields stable tracking but causes performance to plateau earlier. Ultimately, optimizing this framework requires balancing the scaling factor $\lambda$ to prioritize either maximum performance or controller stability.

\section{Conclusion \& Outlook}
\label{sec:conclusion}

This work presents an online-capable, control-informed trajectory planning framework for autonomous racing. Standard software architectures isolate the planning module from downstream execution errors. To solve this, our approach establishes a direct feedback loop from the control module. The framework evaluates control-induced trajectory deviations during runtime and systematically recalculates the spatial constraints of the plannable area. We utilize four mathematical heuristics to compute these boundary adaptations, allowing the MPC planner to safely generate trajectories closer to the true physical limits of the vehicle. Closed-loop simulations confirm that the method improves track utilization and overall lap times while demonstrating online capability.

Future work will explore replacing the heuristic adaptation rules with data-driven algorithms, such as reinforcement learning, to autonomously map complex tracking errors to optimal boundary shifts. Furthermore, integrating online acceleration limit estimation into the observation module could allow the framework to scale the constraint adaptations according to tire degradation or changing environmental conditions. Finally, respecting opponent vehicles on track to use the approach in multi-vehicle scenarios can be investigated. 
\begin{acronym}
  \acro{OCP}{optimal control problem}
\end{acronym}

\end{document}